 \documentclass[final,3p,times,authoryear]{elsarticle}
\usepackage{amssymb}
\usepackage{amsmath}
\usepackage{amsthm}
\newtheorem{definition}{Definition}

\usepackage{subcaption}
\usepackage{xcolor}
\usepackage[ruled,linesnumbered]{algorithm2e}
\usepackage{cleveref}
\usepackage{subcaption}

\journal{Artificial Intelligence in Agriculture}

\begin{document}

\begin{frontmatter}



\title{Interpreting Inflammation Prediction Model via Tag-based Cohort Explanation} %% Article title


\author[ucdavis]{Fanyu Meng}
\author[ucdavis]{Jules Larke}
\author[ucdavis]{Xin Liu}
\author[ucdavis]{Zhaodan Kong}
\author[gatech]{Xin Chen}
\author[ucdavis]{Danielle Lemay}
\author[ucdavis]{Ilias Tagkopoulos}

\affiliation[ucdavis]{organization={University of California, Davis},
            city={Davis},
            state={California},
            country={United States}}
\affiliation[gatech]{organization={Georgia Institute of Technology},
            city={Atlanta},
            state={Georgia},
            country={United States}}

%% Abstract
\begin{abstract}
Machine learning is revolutionizing nutrition science by enabling systems to learn from data and make intelligent decisions. However, the complexity of these models often leads to challenges in understanding their decision-making processes, necessitating the development of explainability techniques to foster trust and increase model transparency. An under-explored type of explanation is cohort explanation, which provides explanations to groups of instances with similar characteristics. Unlike traditional methods that focus on individual explanations or global model behavior, cohort explainability bridges the gap by providing unique insights at an intermediate granularity. We propose a novel framework for identifying cohorts within a dataset based on local feature importance scores, aiming to generate \textit{concise descriptions} of the clusters via tags. We evaluate our framework on a food-based inflammation prediction model and demonstrated that the framework can generate reliable explanations that match domain knowledge.
\end{abstract}

%% Graphical abstract

%% Research highlights
%  \item We introduce a novel approach to cohort explainability with concise cohort descriptions through the use of tags.
%  \item We conduct experiments on a food-based inflammation prediction model and verify that the results match expert knowledge.

%% Keywords
\begin{keyword}
Nutrition Science \sep Inflammation \sep Cohort Explainability \sep Clustering 


%% MSC codes here, in the form: \MSC code \sep code (2000 is the default)

\end{keyword}

\end{frontmatter}


\section{Introduction}
\label{dish}

Machine learning (ML) has become an integral part of modern data analysis, offering powerful tools for uncovering patterns and making predictions across various domains. One significant application is in nutrition science, where ML models can provide dietary recommendations, detect food quality and safety issues during production, and surveil public health and epidemiology. However, the complex and often opaque nature of these models presents challenges in understanding and trusting their predictions. To address these issues, explainability techniques have garnered considerable interest, aiming to make ML models more interpretable and transparent.

Explainability can be approached from different perspectives, including local explanations that focus on individual predictions and global explanations that provide insights into the overall behavior of the model. However, there is a growing need for intermediate-level explanations that balance these two extremes, offering contextually relevant insights that are both comprehensive and specific \citep{fact-sheet, survey2020, survey2018}. Cohort explainability, also referred to as subgroup explainability, explains model predictions by analyzing groups of instances with shared characteristics and emerges as a promising solution to this challenge.

\subsection{Motivating Cohort Explanation}
\label{sec:motiv}

\begin{figure}[htp]
  \centering
  \includegraphics[width=0.3\textwidth]{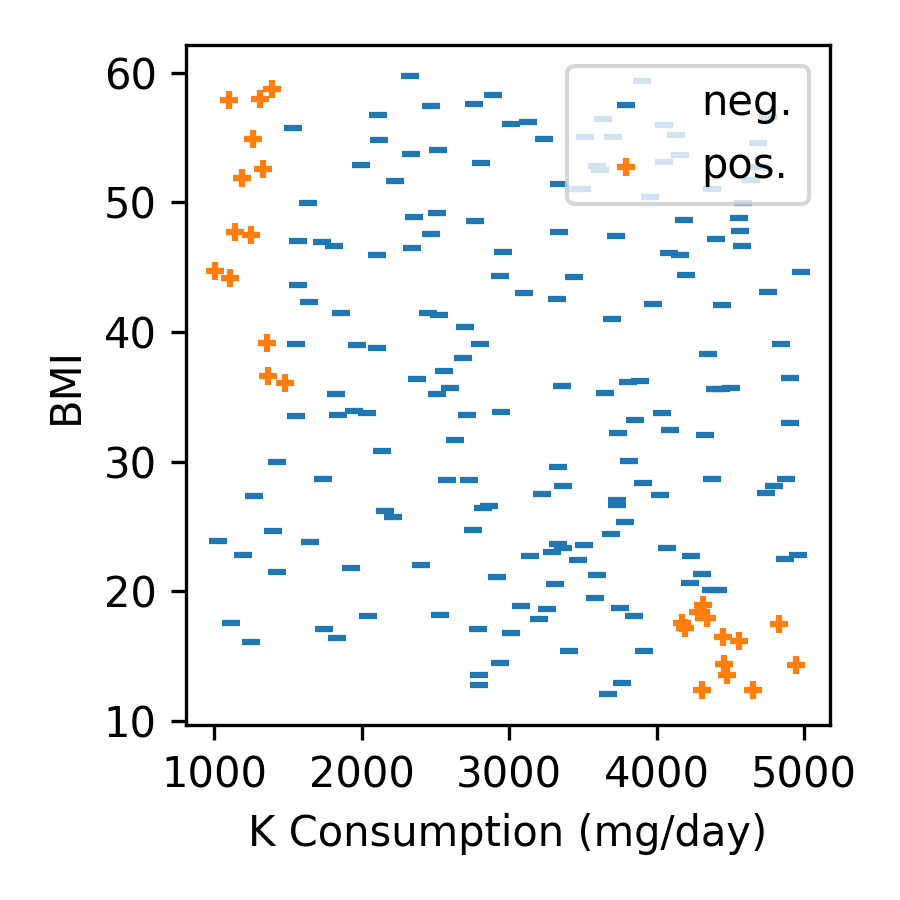}
  \caption{The data distribution of a hypothetical medical disorder classification problem.}
  \label{fig:motiv_data}
\end{figure}

To further motivate the use of cohort explanations, consider a hypothetical medical disorder that predominantly affects low-BMI patients who consume too little potassium, and high-BMI patients who consume too much potassium. The distribution of positive and negative samples is illustrated in Fig.~\ref{fig:motiv_data}. We trained a binary classification model based on two features: BMI and daily potassium consumption. Suppose the model's behavior can be described as the combination of two regional linear models: for patients with low BMI, it assigns a positive weight to potassium and a negative weight to BMI; for patients with high BMI, it assigns a negative weight to potassium and a positive weight to BMI. The model calculates the dot product between the weights and the features and predicts positive if the product is positive.

\begin{figure}[htp]
  \centering
  \begin{subfigure}[t]{0.4\linewidth}
    \centering
    \includegraphics[width=\textwidth]{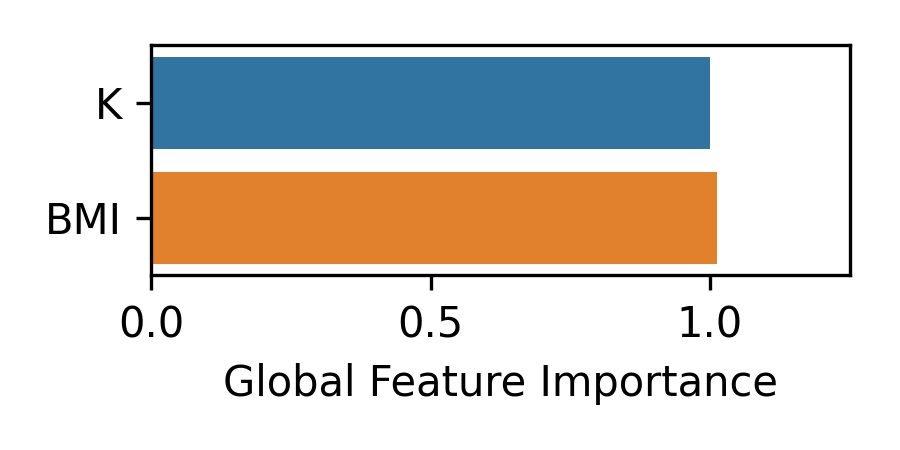}
    \caption{Global feature importance.}
    \label{fig:motiv_global}
  \end{subfigure}\qquad
  \begin{subfigure}[t]{0.4\linewidth}
    \centering
    \includegraphics[width=\textwidth]{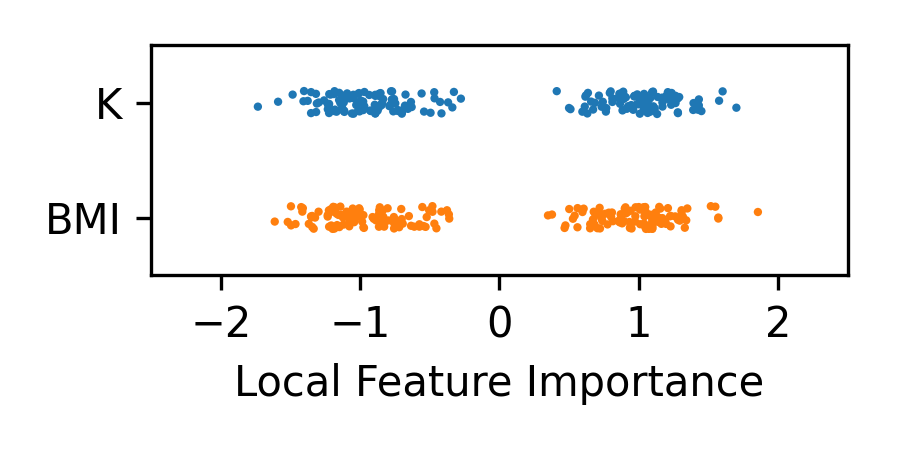}
    \caption{Local feature importance.}
    \label{fig:motiv_local}
  \end{subfigure}
  \begin{subfigure}[t]{0.4\linewidth}
    \centering
    \includegraphics[width=\textwidth]{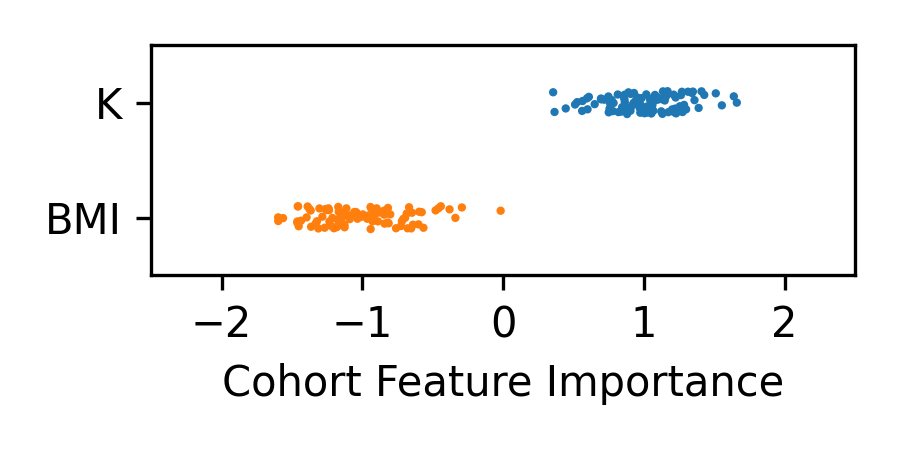}
    \caption{Local importance for low-BMI patients.}
    \label{fig:motiv_c1}
  \end{subfigure}\qquad
  \begin{subfigure}[t]{0.4\linewidth}
    \centering
    \includegraphics[width=\textwidth]{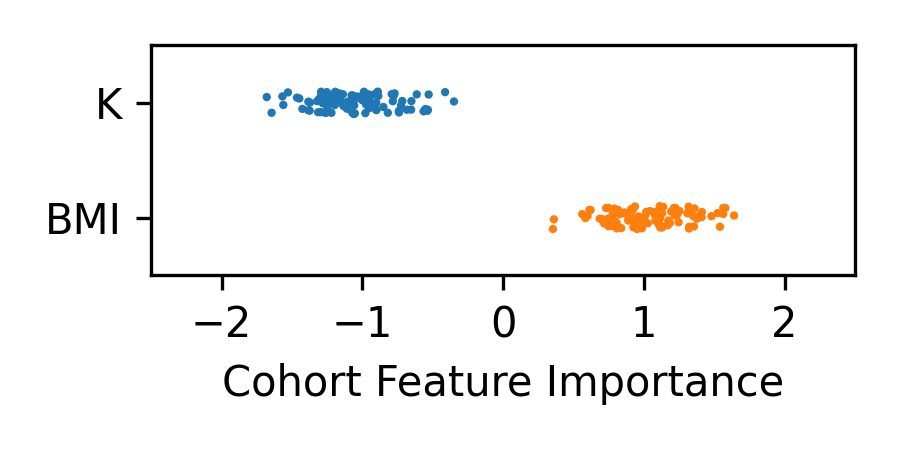}
    \caption{Local importance for high-BMI patients.}
    \label{fig:motiv_c2}
  \end{subfigure}
  \caption{Different types of feature importance explanations for a hypothetical model.}
  \label{fig:motiv}
\end{figure}

If we probe this model using existing local or global explanation methods, the most likely results are shown in Fig.~\ref{fig:motiv_global} and \ref{fig:motiv_local}. In these figures, the feature importance denotes how the machine learning model considers the contribution of each feature to the final prediction, either for an individual sample or averaged over the dataset, respectively. Global importance aims to identify the most important feature to the model. However, since the model treats the features in a completely different manner depending on the patient's age, global importance cannot demonstrate such behavior.

On the other hand, local importance shows how much each feature impacts the model's decision for a particular case. In the figures, a high positive importance on age indicates that, for this sample, the presence of this feature pushes the model's prediction toward a positive outcome. An importance with a low absolute value signifies that this feature contributes little to the final prediction. Local importance can illustrate that the distribution of feature importance may not be uniform across all samples. However, without directly partitioning the samples based on their importance, local importance cannot highlight how feature values correlate with the model's decision-making.

This motivates us to utilize cohort importance, whose goal is to identify subpopulations where the importance of members is similar. The desired cohort explanation for this problem is shown in Fig.~\ref{fig:motiv_c1} and \ref{fig:motiv_c2}. The two cohorts are young and old patients, corresponding to the two regions where the model's behavior differs.

Cohort explanation has significant implications for agriculture and nutrition science. Nutritional recommendations often need to be tailored to specific population groups based on factors such as age, gender, activity level, and existing health conditions. Cohort explanation can assist domain experts with:
\begin{itemize}
  \item \textbf{Simplifying explanations:} Cohort explanation serves as a middle ground between local and global importance methods, offering contextually relevant insights that are both comprehensive and specific. This aggregated view reduces the complexity involved in explaining individual predictions, making it easier for stakeholders to comprehend the model's behavior.
  \item \textbf{Model verification:} Cohort analysis provides a structured approach to evaluate the model's performance across different subsets of samples and can be especially useful with domain knowledge. Typically, this expert knowledge manifests as certain features behaving in a particular way within a specific group of people. By comparing this information with cohort explanation, model engineers can identify regions with issues such as overfitting and underfitting, facilitating targeted sampling and model improvement. For example, in the aforementioned case, if domain knowledge indicates that the condition frequently occurs in high-BMI individuals, regardless of their potassium consumption, then the negative feature importance on potassium in Fig.~\ref{fig:motiv_c2} suggests there are biases in the model, which could potentially be alleviated by re-weighting the dataset to reduce the bias associated with BMI. 
  \item \textbf{Discovering unknown patterns:} Cohort explanation reveals patterns and relationships within specific cohorts, enabling researchers to generate new hypotheses about the underlying mechanisms driving these patterns. For instance, if cohort analysis uncovers a particular dietary factor that significantly impacts certain medical condition levels in a subgroup, this insight can lead to further investigation into the biological or environmental reasons behind this observation.
  \item \textbf{Addressing biases:} By systematically analyzing different cohorts, one can identify and mitigate biases that may exist in the model. Local and global importance can reveal problematic features that the model relies on, while cohort explanation can identify a different type of bias, where the model's behavior changes unfairly across demographic or socioeconomic groups, even though the demographic feature itself may be crucial to the model. This is particularly important in nutrition science and agriculture, where biases in predictions can lead to disparities in recommendations and outcomes.
\end{itemize}

\subsection{How to Describe Cohort Constitution Concisely?}
Cohort explanation can be seen as a middle ground between local explanations, which are specific to individual samples, and global explanations, which provide concise descriptions. Therefore, evaluating cohort explanation methods should take into account both \textit{specificity} and \textit{conciseness}. Most existing work on cohort explainability emphasizes specificity, usually measured by the disparity of local explanations within each cohort. Conciseness, on the other hand, has received less attention and is typically treated as a hyperparameter $k$ representing the desired granularity. Since one of the core motivations of cohort explainability is the balance between these two directions, careful consideration of conciseness is needed. Usually, cohort explanation algorithms describe the cohorts via centroids—a representative sample in the cohort. However, if the feature dimensionality is high, it becomes difficult to comprehend which samples belong to the cohorts. In this work, we propose an alternative: using tags to concisely describe the cohorts.

\subsection{Food-based Inflammation Prediction}
Inflammation associated with poor nutritional status is a strong mediator of many chronic diseases. Current methodologies for assessing diet-induced inflammation in observational studies tend to focus on nutrients and food groups as predictors, which often have limited signals and either capture specific known relationships in the case of nutrients or exhibit broad associations with weaker recommendations. In this work, we use dietary records disaggregated into ingredients and arranged hierarchically to leverage the inherent relationships among foods.

However, even with engineered food intake signals, determining how diet affects inflammation is challenging due to individual-specific characteristics such as age, sex, and body mass, as well as inter-individual variations in genetics, physiology, the host microbiome, and other non-dietary exposures. Cohort explanation directly tackles this problem by providing explanations regarding food effects on automatically identified suitable demographics. Our proposed explainer identifies demographics with similar feature importance. By investigating the composition of the groups and the food effects on each group via feature importance, the explanations could assist nutrition science researchers in further understanding the relationship between food consumption and the causes of inflammation.

In this paper, we focus on post-hoc, feature importance types of explanations such as the commonly used SHAP (SHapley Additive exPlanations) \citep{shap} and LIME (Local Interpretable Model-agnostic Explanations) \citep{lime}.

Our contributions in this paper are as follows:
\begin{enumerate}
  \item We introduce TagHort, a novel approach to cohort explainability with concise cohort descriptions using tags.
  \item We conduct experiments on a food-based inflammation prediction model and verify that the results align with expert knowledge.
\end{enumerate}

\section{Related Work}

\paragraph{Cohort Explanation}

Cohort explainability aims to strike a balance between local and global methods, which provide explanations based on groups of instances. Similar to other explanation types, cohort explanation can be classified into inherent and post-hoc methods. Inherent cohort explanation, sometimes referred to as a local surrogate model, involves dividing the feature space and developing a local, interpretable model for each subspace. Examples of this approach include \citet{slim}, which constructs local-surrogate linear or generalized additive models by recursively partitioning the feature space.

In comparison, post-hoc cohort explanation methods partition the feature space and provide explanations for each cluster. Generally, such explainers run local explanation methods on a dataset and then attempt to cluster the samples based on the similarity between the local explanations. Examples of such methods include \citet{glocal-shap}, which averages local SHAP values within predefined groups. Other methods employ techniques to automatically identify cohorts, such as VINE, which clusters individual conditional expectation (ICE) via unsupervised clustering methods \citep{vine}. Similarly, REPID \citep{repid} and \citet{molnar2023model} use tree-based partitioning to generate cohorts based on either ICE or conditional permutation feature importance. Additionally, GADGET extends previous methods by incorporating functional decomposition during partitioning, allowing for the consideration of more features \citep{gadget}.

The framework proposed in this paper lies within the domain of post-hoc cohort explanation. Compared to existing works, our distinctions include:
\begin{itemize}
  \item We formalize the concept of conciseness in cohort description, which aims to efficiently describe the samples in each cohort.
  \item We introduce descriptive tags to cohort explanation to enhance the conciseness of cohort descriptions.
  \item We utilize dual-view optimization to create cohorts that are both specific and concise.
\end{itemize}

\paragraph{Subgroup Discovery} 

Subgroup discovery is a prominent area in data mining and machine learning that focuses on identifying interesting and interpretable subgroups within a dataset, which is valuable for uncovering hidden structures and actionable insights in various domains \citep{subgroup-survey}. Such works include \citet{sutton2020identifying}, which employs rule-based partitioning based on the loss of surrogate models to identify interesting regions. Similarly, \citet{hedderich2022label} also uses a rule-based algorithm on natural language models to recognize regions with significant responses to certain variables. These methods can be thought of as a type of \textit{inherent} cohort explanation, although the identified clusters of interest may not cover the entire feature space. 

\paragraph{Explainable Clustering}

Explainable clustering is an emerging area in machine learning that aims to enhance the interpretability of clustering results by providing understandable explanations for the derived clusters. For example, \citet{explainable-kmeans} modifies the results from traditional $k$-means into decision trees to improve the interpretability of the clusters. \citet{xclusim} creates a visualization tool to compare and analyze clustering results. More recently, \citet{exp-cluster-optim} formulates a constraint optimization program for clustering while allowing the injection of domain knowledge via constraints and descriptive patterns, which is similar to our formulation. Although the concept of explainable clustering shares similarities with cohort explanation, as both domains utilize explainability and clustering, their goals differ. Cohort explanation aims to apply clustering to simplify the results of local explanations; conversely, explainable clustering seeks to introduce explainability methods to existing clustering algorithms. As for \citet{exp-cluster-optim}, the main difference lies in the target of the optimization problem. \citet{exp-cluster-optim} applies the algorithm to the features, while we focus on clustering the local feature importance scores. Additionally, their main contribution is the introduction of ``discrimination" constraints, which separate the clusters, while our focus is on utilizing as many tags as possible to increase the descriptiveness of the results.

\paragraph{Descriptive Clustering}
Another type of clustering technique is called descriptive clustering, which introduces tags alongside the features. The goal is to find a partition such that, for each cluster, the samples within are close in the feature space while also sharing the maximum number of tags. This approach was first introduced by \citet{desc-cluster0} and has been extended by \citet{desc-cluster1}, \citet{desc-cluster2}, and \citet{desc-cluster3}. This type of dual-objective optimization formulation aligns with the goal of cohort explainability with concise descriptions. In this work, we adapt and simplify the formulation by \citet{desc-cluster0} and apply it to the local feature importance scores instead of the raw features.

\section{Approach}
To address the challenges of specificity and description conciseness simultaneously, we propose a tag-based cohort explanation framework. The algorithm follows these steps:
\begin{enumerate}
  \item Preprocess the features and create a dictionary of tags;
  \item Compute the local feature importance for each sample;
  \item Construct and solve a descriptive clustering optimization problem.
\end{enumerate}

\subsection{Preprocessing}
First, we construct a dictionary of tags that will be used to describe the cohorts. The motivation behind introducing tags is to automate the process of selecting a concise subset of features as descriptions. We first select a subset of "descriptor" features based on the application. We then apply one-hot encoding to categorical descriptor features and discretize continuous features into binary tags. For example, in the motivating example in Fig.~\ref{fig:motiv_data}, we will use both features as descriptors, and the tag dictionary would be \texttt{\{BMI$<$17.5, 17.5$\le$BMI$<$34, 34$\le$BMI$<$51.5, 51.5$\le$BMI, age$<$30, 30$\le$age$<$50, 50$\le$age$<$70, 70$\le$age\}} if we use four quantiles. Each sample will then obtain a binary value for each tag, indicating whether it conforms to the tag. The output of the framework will utilize a subset of these tags to describe each cohort. 

\subsection{Compute Feature Importance}
Next, for the features we wish to investigate, we run a local explainer and compute their importance across all samples. The explainer can be an arbitrary local explainer, including but not limited to model-agnostic methods such as SHAP \citep{shap} or LIME \citep{lime}, or gradient methods for neural network-based models such as saliency maps \citep{saliency}, vanilla gradients, SmoothGrad \citep{smoothgrad}, or Grad$\times$Input \citep{gradxinput}.

\subsection{Descriptive Clustering}
With the binary tags and the local feature importance, we construct an optimization problem. In the following section, we will describe the motivation for the process, while a detailed description of the algorithm can be found in \ref{app:approach}. The goal is to find a partition such that (1) all samples within a cohort conform to a maximum number of common tags; (2) all samples within a cohort have similar local feature importances. 

To solve this problem, we first optimize for maximum \textit{descriptiveness}. Descriptiveness is defined as the minimum number of tags used across the cohorts. This step ensures that each cohort uses as many tags as possible. Note that although we optimize for descriptiveness to enhance the quality of the explanation, the final set of tags will still be much fewer than the total number of tags available.

The optimization problem is subject to the following set of constraints:
\begin{itemize}
  \item Each cohort must have at least one sample;
  \item Each sample must be assigned to exactly one cohort;
  \item A sample can only be placed in a group if it matches all the tags assigned to that group.
  \item A tag can only be assigned to a group \textit{if and only if} every sample in the group fits that tag.
\end{itemize}

Then, among all partitions with good descriptiveness, we aim to find the partition with the best specificity. To achieve this, we construct a similar optimization problem with an additional constraint that descriptiveness cannot decrease. We then solve for the minimum \textit{specificity error}—the sum of pairwise distances of local feature importance within each cohort.

The output of the optimization will be:
\begin{itemize}
  \item The cohort assignment for each sample;
  \item The set of tags used to describe each cohort.
\end{itemize}

Since we introduce specificity as an objective, we enforce each cohort to have similar local feature importances. Thus, we can average the local importance within each cohort to obtain the corresponding cohort explanation.

\subsection{Choosing the Number of Cohorts}
Choosing the granularity of cohort explanation should ideally be handled automatically rather than being estimated by the end users. However, both specificity and descriptiveness tend to increase as the number of cohorts $k$ increases. If $k$ equals the number of samples $n$, then specificity error would be 0, since there would be only one sample in each cohort; and descriptiveness would also be maximized since each sample would utilize the maximum number of tags to describe it. Such results are undesirable, since they degenerate into local explanation and is too complicate for a human to interpret. We prefer a small $k$ with a decently low specificity error.

To automate the process of choosing $k$, we propose a metric \textit{importance prediction error} to evaluate the quality of $k$. The value of $k$ that yields the lowest error is automatically returned. Details of the metric can be found in \ref{app:approach}.

\section{Experiments}

\subsection{Dataset and Explainee Model}

\paragraph{Study Population}

We train an inflammation prediction model based on the National Health and Nutrition Examination Survey (NHANES) 2001-2010 and 2015-2018 cycles, which were selected due to the availability of systemic inflammation (CRP) measurements. 

We apply the following exclusion criteria: participants with diet recalls not passing quality control, caloric intake ranges of $< 500$ or $> 4500$ kcal/day, or those who were below the 5th percentile or above the 95th percentile for that calorie range, individuals under 18 years of age, those with active infections of hepatitis or HIV, diagnoses of cardiovascular disease or cancer, signs of acute inflammation, or a CRP $> 10$ mg/dL, or missing data for covariates. Participants with missing data for BMI were imputed from waist circumference (individually for females and males) using linear models for 197 participants. The final sample size included 19,460 participants after applying the exclusion criteria.

\paragraph{Feature Selection and Prediction Target}

The feature set consists of two components: covariates and diet recalls. Covariates of interest for this study include age, BMI, sex, family poverty income ratio (PIR), education, ethnicity, smoking status, and diagnoses of diabetes and hypertension. Diagnostic criteria to determine diabetic status included any of the following: $\ge 126$ mg/dL fasting blood glucose, $\ge 6.5\%$ glycated blood hemoglobin, or reported use of insulin or other medications. Hypertension was defined as either $\ge 140$ systolic or $\ge 90$ diastolic blood pressure or use of medication for hypertension. Besides the continuous variables age, BMI, and PIR, other discrete covariates were one-hot encoded, forming a total of 20 covariates. 

Foods reported were "ingredientized" to disaggregate mixed meal intake using an automated method. Briefly, the pipeline includes calculating the correct proportion of ingredient weights for items, using text similarity matching with discontinued food in the FNDDS (Food and Nutrition Database for Dietary Studies), and further breakdown of multi-component ingredient descriptions from the Food Commodities Ingredient Database. Dietary data were then processed into a hierarchical "food tree" for feature engineering. Processing diet recalls into the food tree involved binning items that described the same core ingredient into a single clade for further dimensionality reduction. For example, different descriptions of beef would be grouped into a single "Beef, steak" level. Leaf nodes of the food tree (566 leaves) were used as input for an OTU (operational taxonomic unit) table, with grams of intake in dry weight as the abundance measure adjusted for energy intake (per 1000 kcal). Hierarchical feature engineering was employed with TaxaHFE to select the set of dietary features used as input for cohort explanation models to predict the top and bottom tertiles of serum CRP concentrations. Parameters for TaxaHFE included 80 permutations, abundance filter = 0, prevalence filter = 0.1, and lowest level = 2. After applying the feature selection pipeline, we obtained 17 food recall features.

The prediction target, CRP, was divided into a binary target of low and high inflammation based on the lower tertile (N = 6,373) and upper tertile (N = 6,536).

\paragraph{Explainee Model}

We train a binary classifier using XGBoost \citep{xgboost}. We tune the hyperparameters (1) number of estimators; (2) maximum tree depth; (3) row and column subsample ratios; (4) learning rate; (5) leaf threshold $\gamma$; and (6) minimum child weight using 5-fold validation. The model achieves a testing accuracy of 76.4\%.

\subsection{Shapley Additive Explanations (SHAP)}

\begin{figure*}[htp]
  \centering 
  \begin{subfigure}[t]{0.59\textwidth}
    \centering
    \includegraphics[scale=0.09]{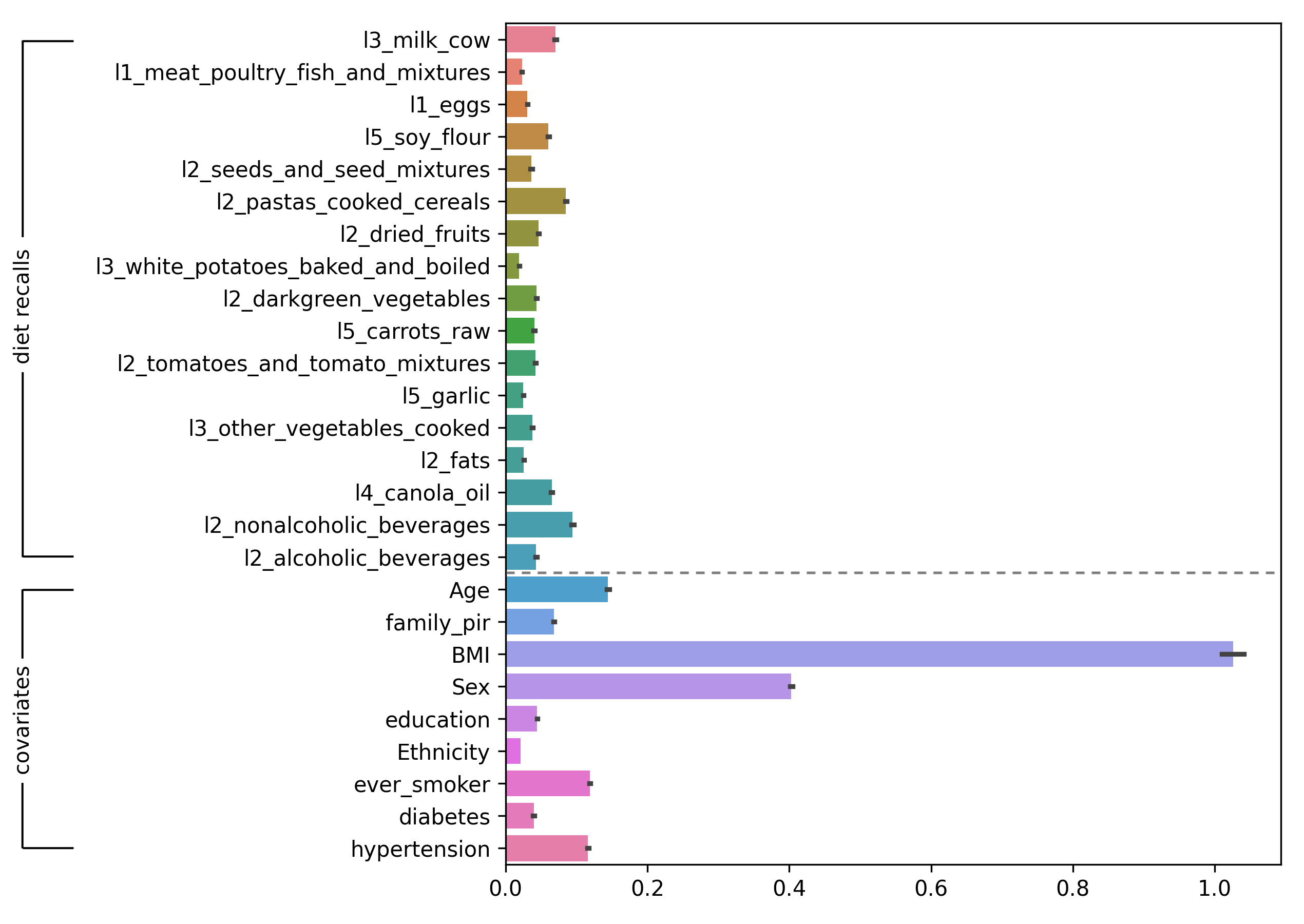}
    \caption{}
    \label{fig:global}
  \end{subfigure}
  \hfill
  \begin{subfigure}[t]{0.39\textwidth}
    \centering
    \includegraphics[scale=0.09]{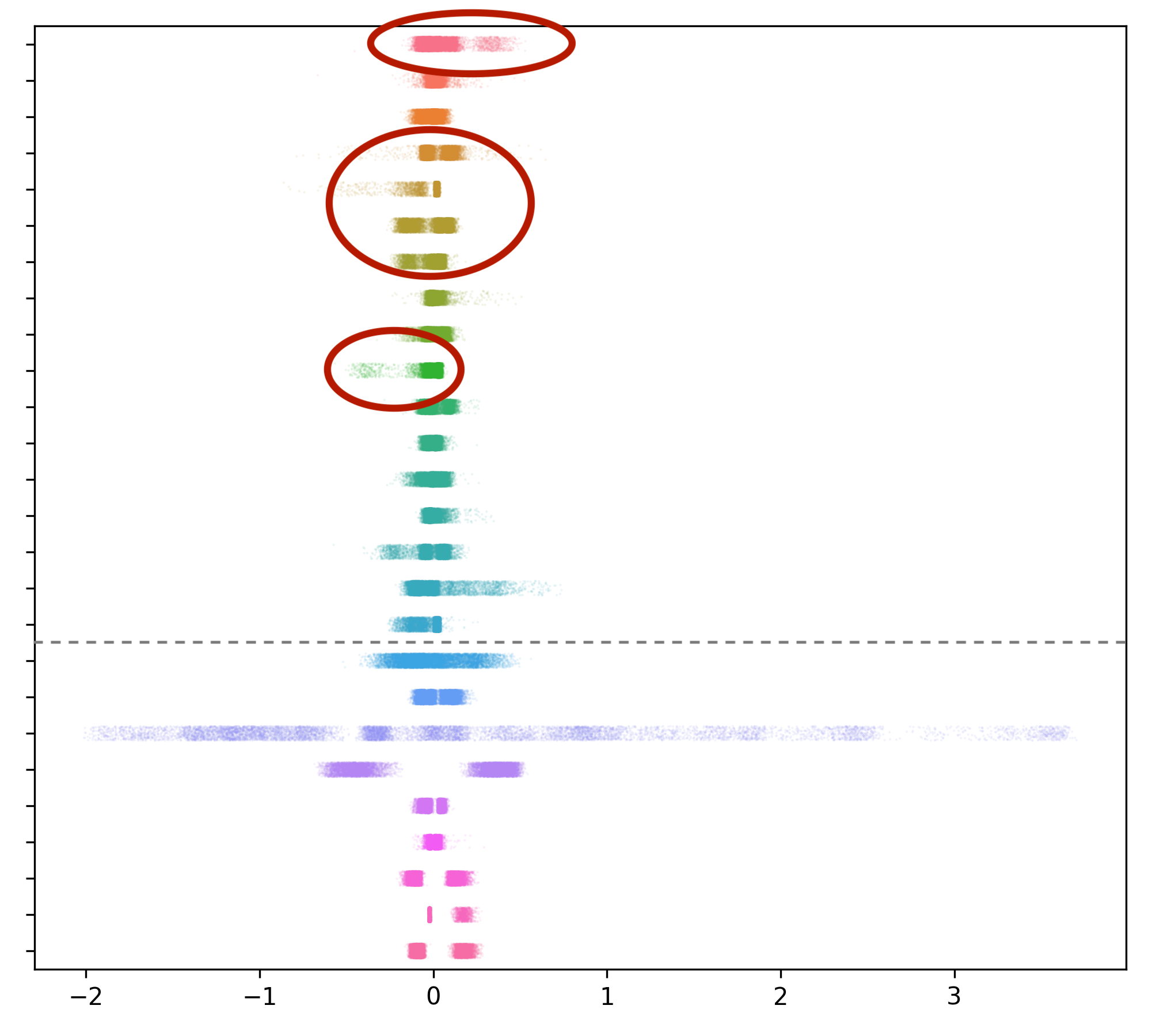}
    \caption{}
    \label{fig:local}
  \end{subfigure}
  \caption{The global and local SHAP importance of the inflammation prediction model. (a) Global importance; (b) distribution of local importance. In (b), we highlight a few food features whose importance distribution appears to have groupings. This motivates the use of cohort explanation to identify the members of those groupings.}
  \label{fig:global-and-local}
\end{figure*}

To increase transparency and interpretability of the model, we first investigate the global and local feature importance using SHAP \citep{shap}. SHAP is a game-theory based method for explaining the output of arbitrary machine learning models. Given a particular sample and its associated prediction, SHAP assigns a value representing the impact on the model output. It calculates this value by comparing the model prediction with and without the feature being present. The SHAP value can be either positive or negative; in our case, a higher positive SHAP value indicates that the feature is positively associated with higher inflammation. Therefore, for the food recalls, we expect that a \textit{high} consumption of pro-inflammatory foods would result in a positive SHAP value, while a \textit{high} consumption of anti-inflammatory foods would yield negative importance. 

We evaluate the SHAP feature importance on the test dataset, which consists of 2,582 samples. Global importance is then computed using the mean absolute value of all local importances. \Cref{fig:global-and-local} shows the results of applying SHAP to the model. In our binary classification task, a higher local feature importance indicates that the feature increases the likelihood of high-level inflammation for that specific individual. We analyze the results as follows:

\begin{itemize}
  \item The model heavily relies on BMI to predict the level of inflammation, which is consistent with domain knowledge indicating that BMI and inflammation are closely correlated \citep{bmi-inflammation};
  \item Other covariates exhibit variable levels of importance to the model, with education and ethnicity having little to no impact. In comparison, food recalls have a small but consistently non-trivial amount of importance;
  \item The distribution of feature importances, including those of the food recalls, is not uniform. This is akin to the motivating example in Sect.~\ref{sec:motiv}, Fig.~\ref{fig:motiv}. Most features exhibit either positive or negative importance depending on the instance. Therefore, the model recognizes that certain foods may have variable impacts on inflammation across different groups of individuals. This further motivates the use of cohort explanation, which can help identify structured groups within the dataset and their corresponding differing effects from food intake. 
\end{itemize}

\subsection{Cohort Explanation}

\paragraph{Cohort Explanation Pipeline} The goal of utilizing cohort explanation is to investigate the varying levels of effect of food intakes on different groups of individuals. Therefore, we focus solely on the importance of food recalls as the importance matrix $W$. This ensures that the optimization algorithm ignores the importance of the covariates, preventing the results from being dominated by BMI, which has a significantly larger importance score. The results of using the importance of all features can be found in the appendix. Regarding the tags $Z$, we transform the covariates into tags. Discrete features such as gender and ethnicity are retained as-is, while we discretize continuous features like age, BMI, and PIR into four quantiles. The optimization problem is solved using MiniZinc \citep{minizinc}. We evaluate the importance prediction error in \Cref{eq:fi-pred-err} across different values of $k$ using 5-fold validation.

\paragraph{Results}

\begin{figure*}[htp]
  \centering
  \begin{subfigure}{0.32\textwidth}
    \centering
    \includegraphics[width=\textwidth]{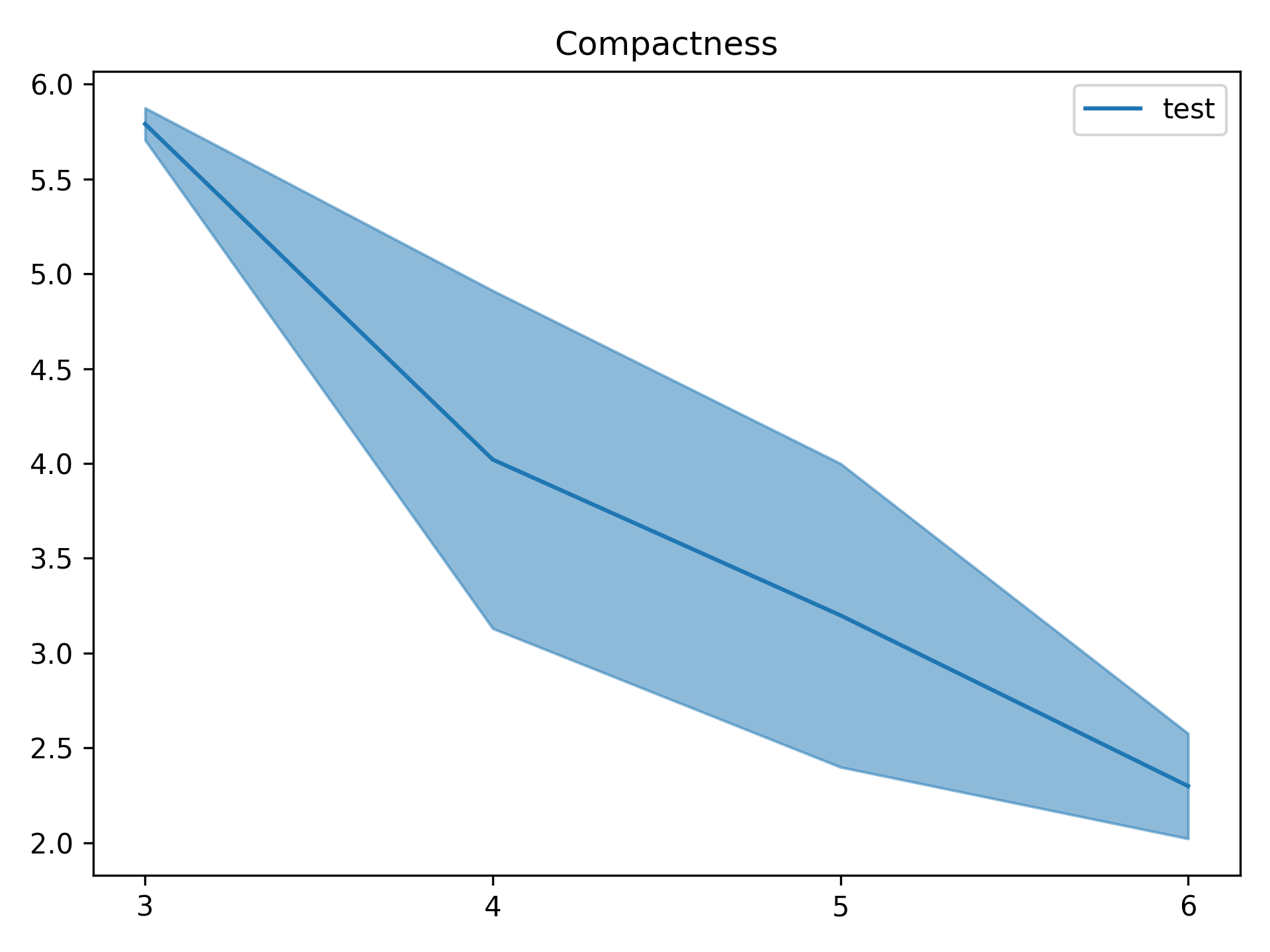}
    \caption{}  
    \label{fig:eval-obj}
  \end{subfigure}
  \begin{subfigure}{0.32\textwidth}
    \centering
    \includegraphics[width=\textwidth]{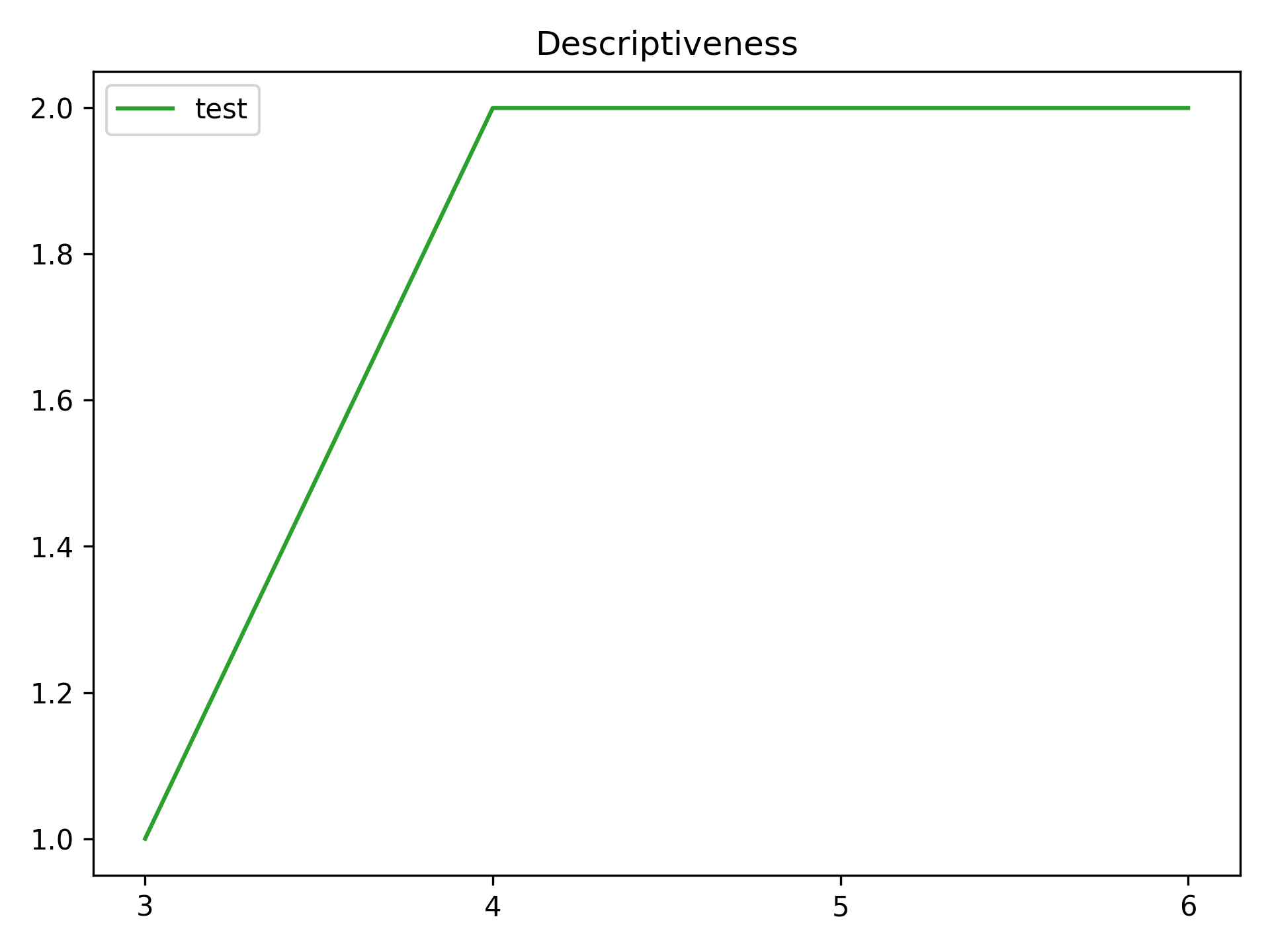}
    \caption{}  
    \label{fig:eval-desc}
  \end{subfigure}
  \begin{subfigure}{0.32\textwidth}
    \centering
    \includegraphics[width=\textwidth]{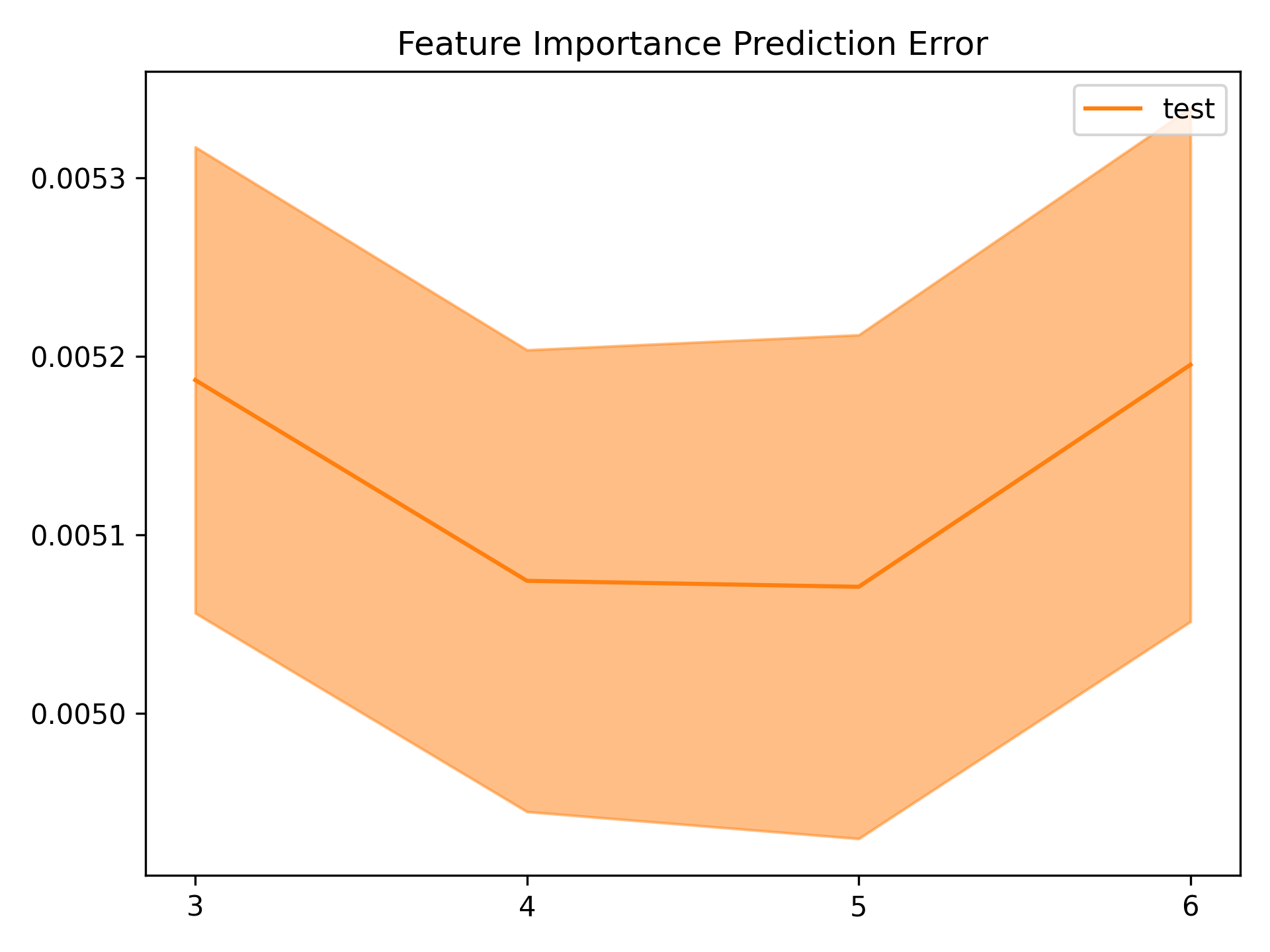}
    \caption{}  
    \label{fig:eval-err}
  \end{subfigure}
  \caption{Evaluating the quality of cohort explanation across different values of $k$. The shaded region depicts the standard deviation. (a) Compactness objective during optimization, which represents the mean inter-cohort pairwise distance on importance, averaged over all cohorts; (b) Descriptiveness objective during optimization, which is the number of tags used by the cohort with the fewest tags; (c) importance prediction error. Note that (b) does not have a shaded region since at all $k$, across all 5 folds, the descriptiveness objective yields the same result.}
  \label{fig:eval}
\end{figure*}

\Cref{fig:eval} illustrates the evaluation metrics at different numbers of cohorts $k$. \Cref{fig:eval-obj,fig:eval-desc} shows the optimization objectives from \Cref{eq:comp,eq:desc}, while \Cref{fig:eval-err} presents the importance prediction error from \Cref{eq:fi-pred-err}. As $k$ increases, both optimization objectives improve; however, descriptiveness plateaus starting from $k=4$ with at least two tags for each cohort. Conversely, the importance prediction error does not improve after $k=4$, indicating that the clustering pipeline is overfitting on the local importance scores. Therefore, we select $k=4$ for our results.

\begin{figure*}[htp]
  \centering
  \includegraphics[width=\textwidth]{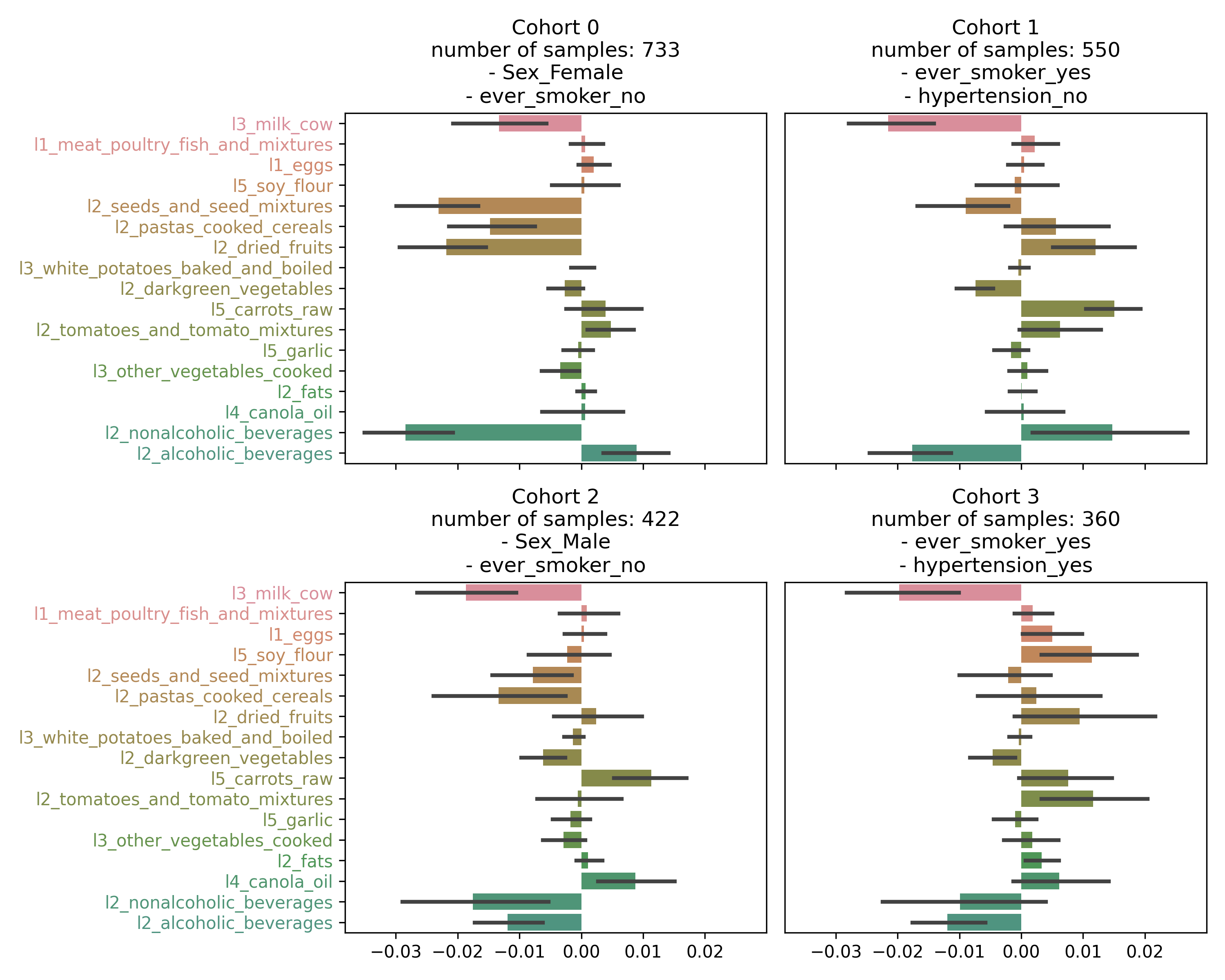}
  \caption{Cohort importance and their descriptions using TagHort with $k=4$. The title of each subplot also includes the tags that describe the cohort.}
  \label{fig:k=4-abs}
\end{figure*}

\begin{figure*}[htp]
  \centering
  \includegraphics[width=\textwidth]{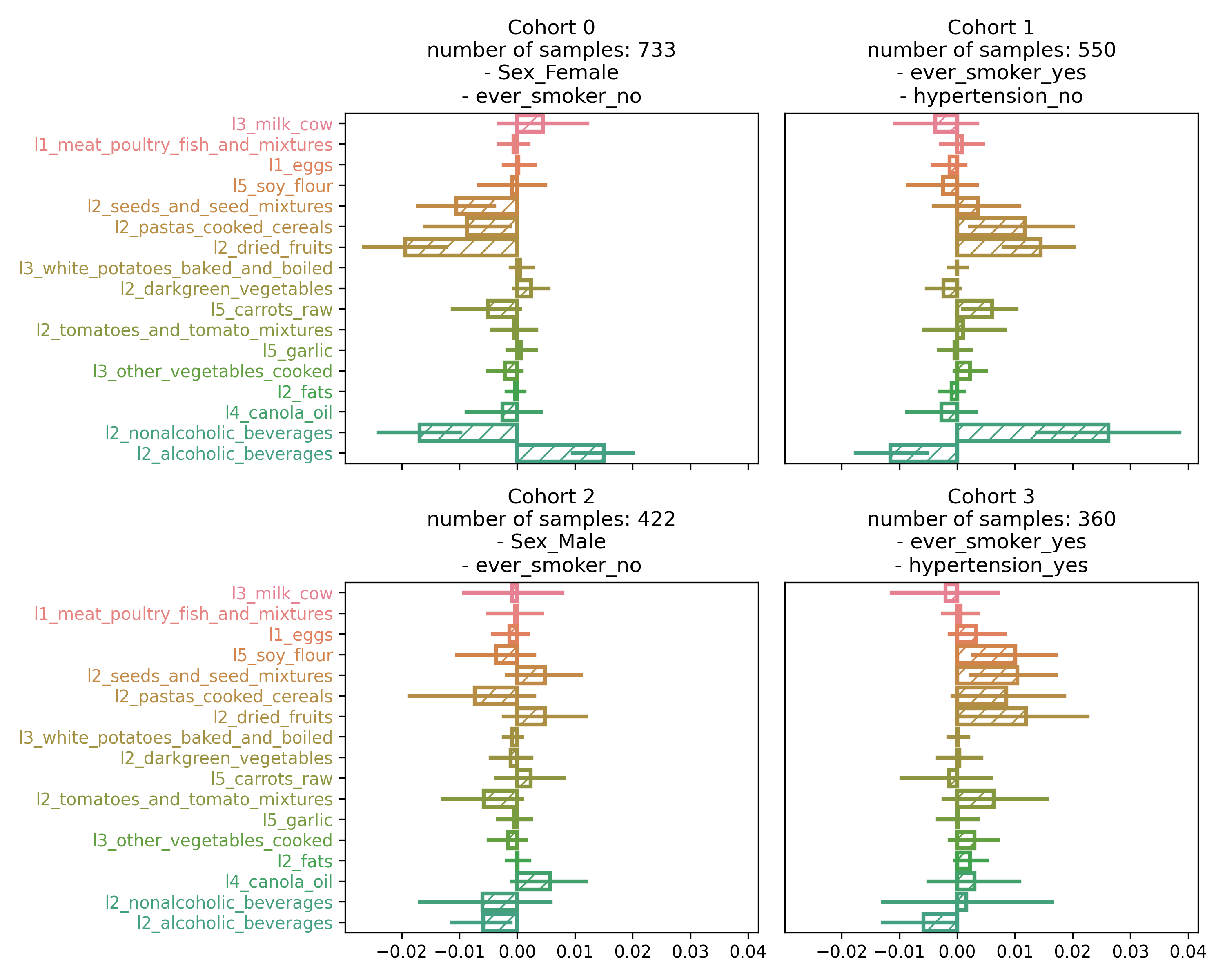}
  \caption{Relative cohort importance and their descriptions using TagHort with $k=4$. The title of each subplot also includes the tags that describe the cohort.}
  \label{fig:k=4-diff}
\end{figure*}

\Cref{fig:k=4-abs,fig:k=4-diff} displays the results for $k=4$. For each cohort, we compute cohort importance as the mean local importance for samples belonging to that cohort. \Cref{fig:k=4-abs} illustrates the mean importance within each cohort, while \Cref{fig:k=4-diff} shows the \textit{difference} between the cohort mean and the dataset mean importance. Thus, in \Cref{fig:k=4-diff}, a positive value indicates that, within this cohort, the feature importance of the specific food is associated with greater inflammation or is less effective in reducing inflammation. We reiterate that the cohorts may not be linearly partitioned in the feature space. That is, although all samples in cohort 0 exhibit the properties "sex = female" and "ever\_smoker = no," it does not guarantee that all female non-smokers belong to cohort 0.

We note the following observations from the cohort explanations:
\begin{itemize}
  \item Both cohort 1 and 3 contain the tag "ever\_smoking = yes." Both cohorts show a higher-than-average importance for seeds, dried fruits, carrots, and tomatoes. This aligns with findings that smokers require additional vitamin C to meet their needs \citep{smoking-vc}. The model recognizes this and concludes that fruits and vegetables are less effective in reducing inflammation in smokers due to heightened baseline inflammation caused by smoking;
  \item Similarly, in the two smoker cohorts, non-alcoholic beverages exhibit relatively higher importance compared to those who do not smoke. This category includes sodas and fruit beverages, which have high sugar content and are known to induce inflammation. This disparity aligns with research suggesting that smoking and sugar may jointly contribute to inflammation \citep{smoking-soda};
  \item Within smokers, non-alcoholic beverages are assigned pro-inflammatory importance more significantly among individuals \textit{without} hypertension. One possible explanation is that sugary beverages are correlated with inflammation \citep{hypertension-soda-1, hypertension-soda-2}. Since hypertension and inflammation are often linked \citep{hypertension-inflammation}, the model finds the effect of such beverages much more prominent in individuals without hypertension;
  \item Other significant differences include the model considering alcoholic beverages to be most pro-inflammatory among female non-smokers. Such distinctions could potentially indicate that the model recognizes hidden confounders within female non-smokers or that it has biases regarding this subpopulation.
\end{itemize}

\begin{figure*}[htp]
  \centering
  \begin{subfigure}{0.45\textwidth}
    \centering
    \includegraphics[width=\textwidth]{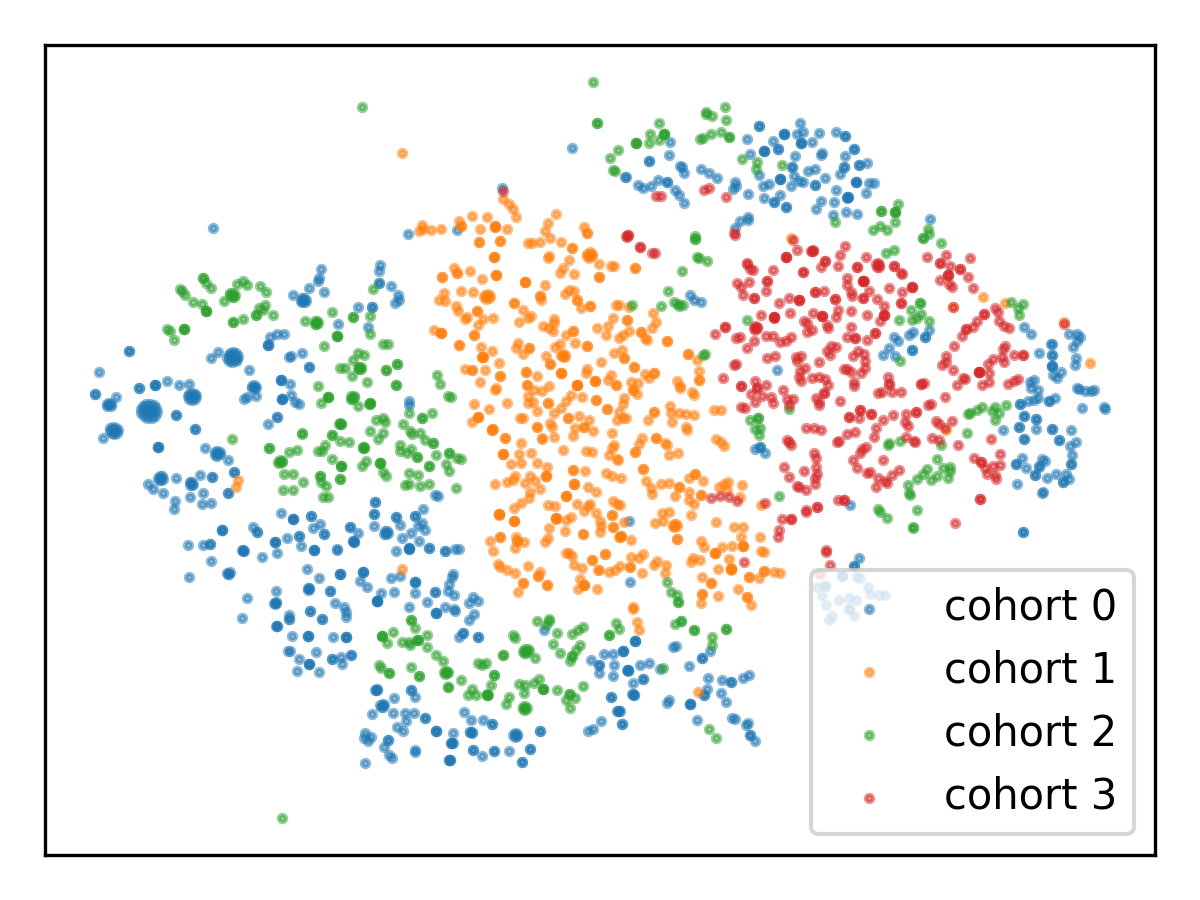}
    \caption{Feature space.}
  \end{subfigure}
  \begin{subfigure}{0.45\textwidth}
    \centering
    \includegraphics[width=\textwidth]{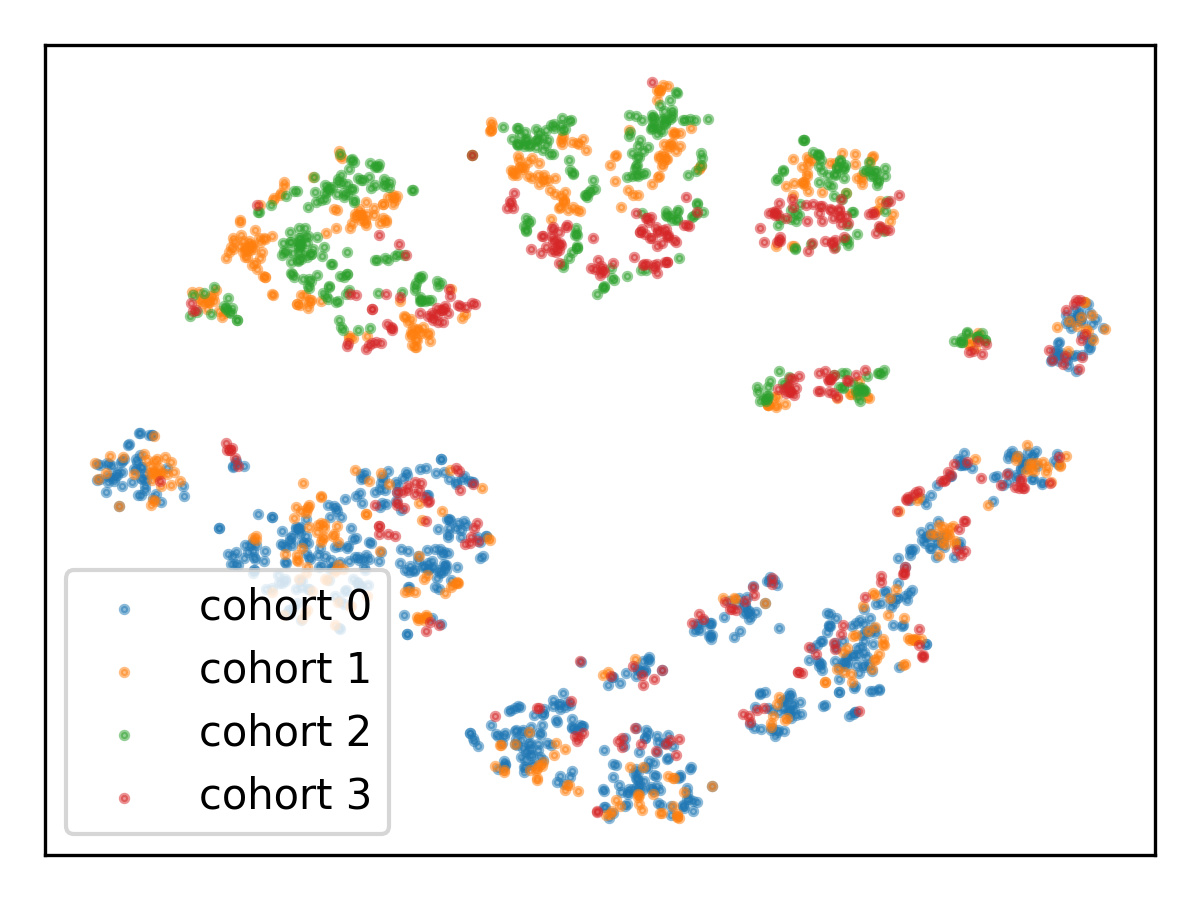}
    \caption{SHAP local importance space.}
  \end{subfigure}
  \caption{Visualizing the cohorts from TagHort in the feature space and the SHAP local importance space. The figure is obtained by applying T-SNE to either the covariates or the \textit{importance} of food recalls. The same color indicates that the samples belong to the same cohort.}
  \label{fig:tsne}
\end{figure*}

Additionally, \Cref{fig:tsne} visualizes the partitioning of the identified cohorts based on both the features and the importance scores. Samples belonging to the same cohort are generally close both in their feature values and in their local importance scores. This demonstrates that the pipeline is capable of finding structured partitioning in both dimensions.

In conclusion, our tag-based cohort explanation effectively identifies interesting groupings within the dataset, and the explanations are consistent with current research findings.

\subsection{Comparison with REPID}

To compare our framework with existing literature on cohort explanation, we selected REPID \citep{repid}, which applies a decision tree to the local feature importance. Since REPID utilizes the inherently interpretable decision tree for clustering, its descriptions of the cohorts are concise compared to centroid-based methods such as VINE \citep{vine} and are on par with our framework.

\begin{figure}[htp]
  \centering
  \includegraphics[width=0.5\textwidth]{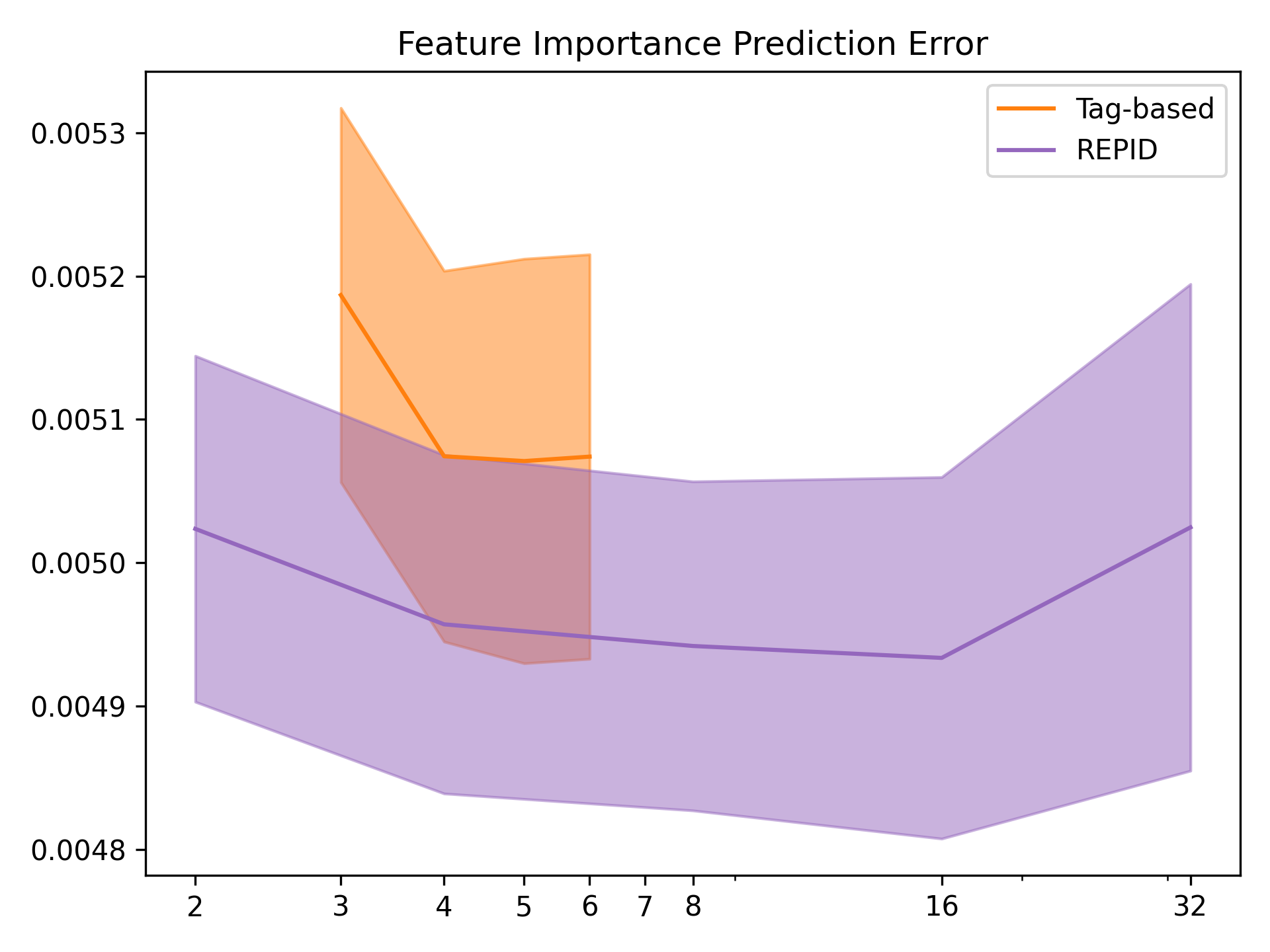}
  \caption{Comparison of importance prediction error between the proposed TagHort and REPID across different values of $k$.}
  \label{fig:eval-cmp}
\end{figure}

\begin{figure}[htp]
  \centering
  \includegraphics[width=\textwidth]{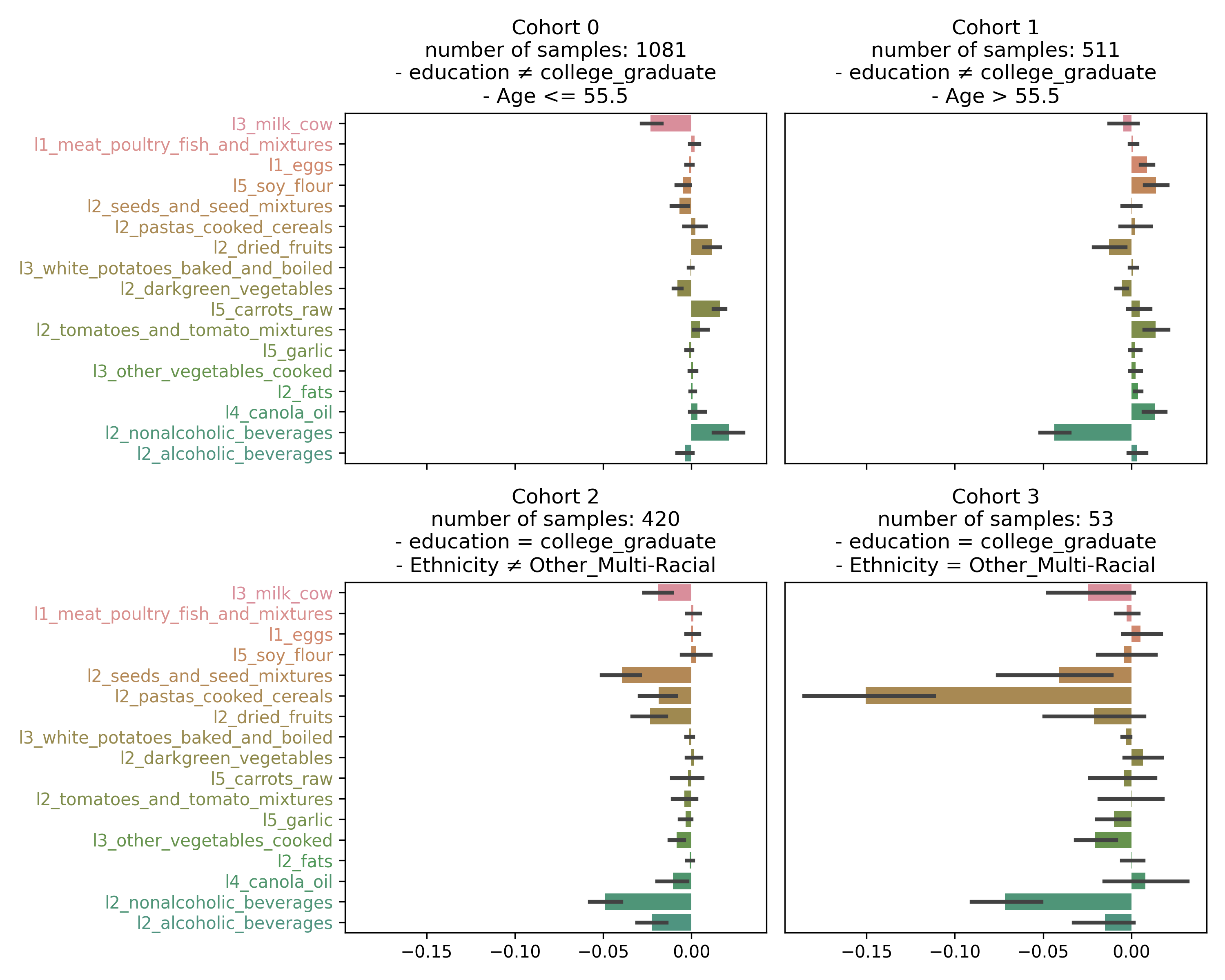}
  \caption{Cohort importance and their descriptions using REPID at $k=4$. The title of each subplot also includes the tags that describe the cohort.}
  \label{fig:repid}
\end{figure}

\Cref{fig:eval-cmp} shows the evaluation based on importance prediction error between the proposed TagHort and REPID across different values of $k$, while \Cref{fig:repid} presents the explanation generated by REPID at $k=4$. REPID consistently achieves better importance prediction error, which is to be expected since descriptive cohort explanation does not guarantee disjoint cohorts. Nevertheless, \Cref{fig:repid} illustrates that without such a method, the results may not yield an explainable partition and/or explanation. Due to the high dimensionality of the features, the decision tree does not perform well and produces imbalanced cohorts. Consequently, it focuses on features such as ethnicity and education, which are less effective in identifying importance-based cohorts. 

\begin{figure}[htp]
  \centering
  \includegraphics[width=\textwidth]{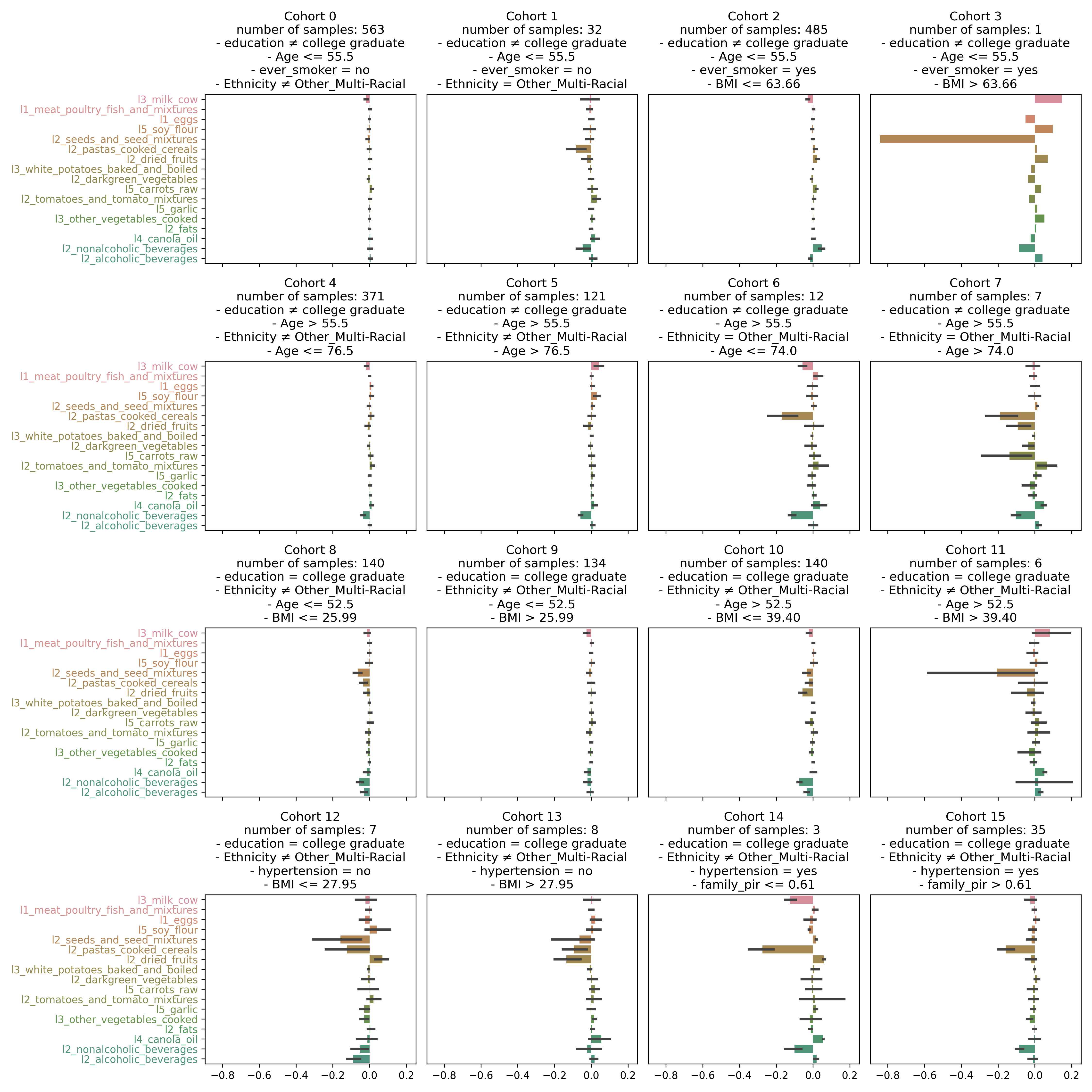}
  \caption{Cohort importance and their descriptions using REPID at $k=16$. The title of each subplot also includes the tags that describe the cohort.}
  \label{fig:repid-16}
\end{figure}

We also evaluate REPID at $k=16$, where it achieves the best importance prediction error, which can be found in \Cref{fig:repid-16}. Although REPID now introduces more useful features such as age, BMI, and hypertension as descriptors, the cohorts remain imbalanced. More than half of the cohorts contain fewer than 40 samples, and comparing results between cohorts of vastly different sizes is generally not useful. This demonstrates that traditional clustering schemes do not perform well under high dimensionality.

\section{Conclusion}

In this paper, we introduced TagHort, a novel framework for tag-based cohort explanation with concise descriptions using tags. Our approach bridges the gap between local and global explainability by providing intermediate-level insights that are both specific and comprehensive. We demonstrated the effectiveness of our framework on a food-based inflammation prediction model, showing that it can generate reliable explanations consistent with domain knowledge. Through our experiments, we illustrated the importance of cohort explanation in identifying structured groups within datasets and understanding the varying effects of features across different populations. Our method provides a valuable tool for enhancing model transparency and trust, with significant implications for healthcare and nutrition science. Future work could explore the integration of additional types of tags and further validation across different domains and models.

\bibliographystyle{elsarticle-harv} 
\bibliography{taghort.bib}

\clearpage
\appendix
\section{Details on the Tag-based Cohort Explanation Framework}
\label{app:approach}

\subsection{Problem Definition}

Consider a fixed, black-box model $M: \mathbb{R}^m \mapsto \mathbb{R}^q$, and a local feature importance explainer used to explain the model $\omega_M: \mathbb{R}^m \mapsto \mathbb{R}^m$. We are given samples $X = \{x_1, \dots, x_n\}, x_i \in \mathbb{R}^{m}$ and can thus evaluate the local importance score of each of the $n$ samples via $\omega_M$ to obtain $W = \{w_1, \dots, w_n\}, w_i \in \mathbb{R}^{m}$. Additionally, for each sample, we have $p$ auxiliary binary tags that describe the sample $D \in \{0, 1\}^{n \times r}$. In our application, these tags are obtained from the features via preprocessing.

\begin{definition}[Tag-based cohort explanation]
  Given the desired number of cohorts $k$, the goal of cohort explanation with concise description is to find a partition $C = \{C_1, \dots, C_k\}$ such that the local importance scores are compact, i.e.,
  \begin{equation}
    \mathrm{compactness}(C) = 
    \sum_{t=1}^k \sum_{x_i, x_j \in C_t, i \neq j} \|w_i - w_j\|_2^2
    \label{eq:comp} 
  \end{equation}
  is minimized.
  
  Additionally, each cohort $C_t$ should provide a list of tags $S_t$ that describe the cohort, i.e.,
  \begin{equation}
    x_i \in C_t \Longrightarrow D_{is} = 1, \quad \forall s \in S_t
  \end{equation}
  
  Ideally, we would like the result to utilize as many tags as possible, and thus the quality of $\mathcal{C}$ is also evaluated by
  \begin{equation}
    \mathrm{descriptiveness}(C) = \min_{t=1}^k |S_t|
    \label{eq:desc}
  \end{equation} 
  where $|S_t|$ represents the number of tags used to describe cohort $t$.
\end{definition}

\subsection{Formulation}
To solve the problem, we simplify the formulation proposed by \citet{desc-cluster0} and apply it to the local importance scores $\mathcal{W}$. We model the problem as a constraint problem (CP).

\subsubsection{Variables}

We introduce two variables:

\begin{enumerate}
  \item Cohort assignment $G \in \{1, \dots, k\}^n$, where $G_i$ represents the index of the cohort to which $x_i$ belongs.
  \item Description matrix $S \in \{0, 1\}^{k\times r}$, where $S_{tp} = 1$ indicates that the $p$-th tag is used to describe cohort $t$.
\end{enumerate}

\subsubsection{Constraints}
The constraints for the CP are as follows:
\begin{itemize}
  \item Each cohort should be non-empty:
    \begin{equation}
      \mathrm{count}(G, t) >= 1, \quad \forall t = 1, \dots k
      \label{eq:cons-ne}
    \end{equation}
    where $\mathrm{count}(G, t)$ denotes the number of occurrences of the value $t$ in $G$;
  \item A sample can be placed into a cohort if it satisfies all of its tags:
    \begin{equation}
      \sum_{p=1}^r S_{G_i,p} (1 - D_{ip}) = 0, \quad \forall i = 1, \dots, n
      \label{eq:cons-sam}
    \end{equation}
  \item A tag should be used to describe a cohort if all samples in it satisfy the tag:
    \begin{equation}
      S_{tp} = 1 \Longleftrightarrow \sum_{i=1, G_i = t}^n (1 - D_{ip}) = 0
      \label{eq:cons-tag}
    \end{equation}
  \item Tie-breaking:
    \begin{equation}
      \mathrm{seq\_precede\_chain}(G)
      \label{eq:cons-tie}
    \end{equation}
    which is a global constraint that enforces the value $t$ to precede all values of $t+1$ in $G$. This allows earlier samples to be put in earlier groups, and thus prevent ties.
\end{itemize}

\subsection{Tag-based Cohort Explanation Framework}

\begin{algorithm}[htp]
\small
\caption{Tag-based Cohort Explanation}
\label{alg:taghort}
\SetKwInOut{Input}{Input}
\SetKwInOut{Output}{Output}
\Input{Local importance $W$, tags $D$, number of cohorts $k$}
\Output{Cohort assignment $G$, description matrix $S$}
$\mathcal{C} \gets $ constraints defined in Eq.~\ref{eq:cons-ne}-\ref{eq:cons-tie} \;
$q \gets \max \mathrm{descriptiveness}(C)$ subject to $\mathcal{C}$ \;
$G, S \gets \arg \min \mathrm{compactness}(C)$ subject to $\mathcal{C} \cup \{\mathrm{descriptiveness}(C) \ge q \}$ \;
\For{$t \gets 1$ \textbf{to} $k$} {
  $\bar{w}_t \gets \mathrm{mean}(\{W_i | G_i = k\})$
}
\Return $G, S, \{\bar{w}_t\}_{t=1}^k$\;
\end{algorithm}

\Cref{alg:taghort} describes the workflow of our proposed tag-based explanation framework. We construct two CPs, one for descriptiveness in Eq.~\ref{eq:desc} and one for compactness in Eq.~\ref{eq:comp}. We first find the optimal descriptiveness of the partitions by maximizing tag usage. Then, we solve for compactness while enforcing descriptiveness to not decrease. This process allows us to find a clustering that is both compact and descriptive. 

Unlike \citet{desc-cluster0}, we do not repeatedly solve for the two objectives to obtain a full Pareto front. Instead, we only run one pass and solve each problem once. Since cohort explainability highly values the quality of the descriptions, a partition with fewer utilized tags is undesirable. Therefore, we only optimize for compactness under the optimal descriptiveness.

Additionally, unlike existing cohort explanations such as VINE or REPID, the cohort generated by the proposed method may not be disjoint in terms of features or tags. Two cohorts can share some of their tags, and it is possible for a new unseen sample to satisfy the tag requirements of multiple cohorts. We interpret this as our method identifying \textit{necessary} cohorts, while most existing literature provides \textit{sufficient} cohorts. In other words, our explanation template is that ``for a sample $x$ to have importance similar to $\bar{w}_t$, it should satisfy the descriptions of cohort $t$," whereas disjoint clustering offers explanations in the form of ``if a sample $x$ is assigned to cohort $t$, its importance should be similar to $\bar{w}_t$."

\subsection{Choosing the Hyperparameter $k$}

Choosing the granularity of cohort explanation should ideally be automated rather than guessed by end users. However, the two primary objectives of tag-based explanation, descriptiveness and compactness, both improve as $k$ increases. Consequently, we propose a metric, \textit{importance prediction error}, to evaluate the quality of a specific $k$.

\begin{definition}[Importance prediction error]
  Given cohorts $C_1, \dots, C_k$ and description matrix $S$, along with an evaluation dataset $X' \in \mathbb{R}^{l\times m}$ and their corresponding local importance scores $W' \in \mathbb{R}^{l\times m}$ and tags $D'\in \{0,1\}^{l\times r}$, the importance prediction error is defined as
  \begin{equation}
    \mathcal{L}_\mathrm{FI} = \sum_{i=1}^l \left\|w'_i - \frac{1}{|Z_i|} \sum_{C \in Z_i} \bar{w}_t\right\|_2^2
    \label{eq:fi-pred-err}
  \end{equation}
  where $Z_i$ is the set of all cohorts such that sample $x'_i$ satisfies all of their descriptions:
  \begin{equation}
    C_t \in Z_i \Longleftrightarrow \left( S_{tp} = 1 \Longrightarrow D'_{ip} = 1, \forall p = 1, \dots, r \right)
  \end{equation}
\end{definition}

In essence, we treat cohort explainability as a supervised learning problem, where we aim to predict the local importance score based on its tags. If a sample matches multiple cohorts, we predict its importance as the average of the importances of all the cohorts it satisfies.

Note that this notion of importance prediction error is also applicable to other types of cohort explainability methods such as REPID. In such cases, we do not need to average the importance of multiple cohorts since all samples must belong to exactly one cohort according to their partitioning scheme.

\end{document}